\documentclass[11pt]{article}
\usepackage[preprint]{acl}
\usepackage{times}
\usepackage{latexsym}
\usepackage[T1]{fontenc}
\usepackage[utf8]{inputenc}
\usepackage{microtype}
\usepackage{inconsolata}
\usepackage{graphicx}
\usepackage{amsmath}
\usepackage{amssymb}
\usepackage{booktabs}
\usepackage{multirow}
\usepackage{algorithm}
\usepackage{algpseudocode}
\usepackage{adjustbox}
\usepackage{enumitem}
\newcommand{\includefigure}[2][]{\includegraphics[#1]{#2}}

\title{EchoKV: Self-Corrective KV Cache Eviction\\
       via Iterative Query Feedback}

\author{JianKun Zou \quad Tongtong Yuan \\
  Beijing University of Technology \\
  \texttt{zoujiankun2006@emails.bjut.edu.cn} \\
  \texttt{yuantt@bjut.edu.cn}}

\hypersetup{pdftitle={EchoKV: Self-Corrective KV Cache Eviction via Iterative Query Feedback},pdfauthor={JianKun Zou and Tongtong Yuan}}

\begin{document}
\maketitle

\begin{abstract}
KV cache eviction is not merely a token-ranking problem: the workspace retained
for generating eviction queries also determines the quality of those queries.
Consequently, an early workspace error can distort later query observations and
reinforce an incorrect final eviction decision, particularly when evidence is
dispersed across a long context.
We present \textbf{EchoKV}, which discovers and addresses this failure mode
with iterative correction. It initializes a feedback-compatible bootstrap
workspace using proposals at future decode positions, then uses early
decode-generated queries to recover the workspace before the remaining
lookahead steps. It uses the same number of workspace forward passes as LAQ
while adding a tensor re-scoring operation for workspace repair.
On LongBench (16 tasks, Mistral-7B-Instruct-v0.2 and Qwen2.5-7B-Instruct),
EchoKV shows notable improvements over LAQ on several tasks requiring
dispersed evidence, including multi-hop cross-document QA and long
single-document comprehension.
Across budgets and backbones, EchoKV selectively improves evidence-intensive
tasks while maintaining comparable performance on the remaining LongBench
subsets. These results identify query--workspace coupling as a practical
failure mode of query-driven KV eviction.
\end{abstract}

\section{Introduction}

KV cache eviction is not merely a token-ranking problem. In query-driven
eviction, the tokens retained during workspace construction determine the
future queries used for subsequent eviction decisions. This creates a
self-dependent feedback loop: an early omission changes the hidden states,
the changed hidden states produce drifting queries, and those queries can
reinforce the original omission.

The deployment of LLMs in long-context settings is bottlenecked by the
growth of the KV cache.
Eviction-based approaches~\citep{zhang2023h2o,li2024snapkv,wang2025lookahead}
selectively discard cache entries deemed unimportant for future generation,
enabling sub-linear cache sizes at the cost of approximation error.

A key insight of LAQ~\citep{wang2025lookahead} is that eviction quality
improves when the selection query reflects \emph{future} decode queries rather
than the tail of the prompt.
LAQ generates $T$ greedy lookahead tokens on a compressed workspace, collects
their queries, and uses them to re-score the full prompt KV.
While effective, LAQ leaves this feedback loop unresolved:

\paragraph{Window-biased stage-1 compression.}
LAQ builds its lookahead workspace via SnapKV~\citep{li2024snapkv}, which
scores token importance using only the last $w$ prompt queries.
For tasks where relevant content is spread across multiple document sections
(e.g.\ \textsc{Qasper}) or requires multi-hop reasoning
(e.g.\ \textsc{HotpotQA}), the window is unlikely to capture all necessary
evidence, leading to a degraded workspace and, consequently, degraded
lookahead queries.

This failure is self-reinforcing rather than a one-time ranking error. Once an
initial workspace excludes relevant evidence, the model generates lookahead
queries from hidden states that cannot attend to that evidence; later eviction
scores are therefore conditioned on a distorted observation. A robust solution
must address both stages of this loop: it must construct a decoding-aligned
workspace before any generated query is available, and it must give the model
an opportunity to use early queries to revise an imperfect initial workspace.
This perspective motivates EchoKV's feedback-compatible workspace bootstrap
and its feedback-driven workspace repair.

We present \textbf{EchoKV} with the following contributions:
\begin{enumerate}
    \item \textbf{Diagnosis.}
          We formulate query-driven KV eviction as a partially observable,
          self-dependent process: workspace quality conditions the query
          observations used to make the next eviction decision. We provide a
          direct measurement (Section~\ref{sec:obs}) showing that query
          fidelity to true decode queries degrades with lookahead depth on a
          frozen workspace.
    \item \textbf{Closed-loop correction.}
          EchoKV turns KV eviction into a closed-loop correction process. It
          constructs a \emph{feedback-compatible bootstrap workspace} before
          any decoded query exists, then uses round-1 generated queries not
          only for final scoring but also for \emph{feedback-driven workspace
          repair}. Future-position proposals are used only for initialization.
          The method uses $T$ forward passes over a $B_1$-sized workspace---
          identical to LAQ---plus one tensor re-scoring pass.
    \item \textbf{Targeted gains without overclaiming.}
          On 16-task LongBench, EchoKV's most consistent gains occur on
          dispersed-evidence tasks such as multi-hop QA and cross-passage
          retrieval. Its aggregate performance is comparable to LAQ, which
          motivates a failure-mode-specific claim rather
          than a universal one.
\end{enumerate}

\section{Related Work}

\paragraph{KV cache compression.}
Context-extension and efficient-attention methods
\citep{chen2023extending,peng2024yarn,ding2024longrope,chen2024longlora,
dao2022flashattention,beltagy2020longformer} enlarge or accelerate attention,
but do not select which KV entries remain stored. EchoKV is complementary:
it operates at inference time on an unmodified decoder.
StreamingLLM~\citep{xiao2023streamingllm} retains only sink and recent tokens.
H2O~\citep{zhang2023h2o} keeps globally heavy-hitter tokens via accumulated attention.
SnapKV~\citep{li2024snapkv} uses the attention of the prompt's final window to score tokens.
PyramidKV~\citep{cai2024pyramidkv} allocates layer-varying budgets.
AdaKV~\citep{feng2024adakv} and HeadKV~\citep{fu2024headkv} further differentiate budgets per attention head.
All these methods rely on prompt-phase queries for eviction, ignoring decode-phase query directions.
Orthogonal approaches reduce the byte footprint of every retained KV entry
rather than selecting a subset of tokens. KIVI~\citep{liu2024kivi} uses
asymmetric low-bit quantization for keys and values, while
KVQuant~\citep{hooper2024kvquant} combines low-precision quantization with
outlier handling. These methods are complementary to EchoKV: quantization
changes the representation precision, whereas EchoKV improves which entries
are retained for the decode cache.

\paragraph{Pseudo-query eviction.}
LAQ~\citep{wang2025lookahead} first introduced greedy lookahead queries for KV eviction
and demonstrated consistent gains over window-based methods across budgets.
DapQ~\citep{dapq2026} shows that query position can strongly affect which keys
are ranked highest, and therefore replaces the observation window with
position-aware pseudo-queries synthesised at future decode positions---obtaining
a decoding-aligned observation window without decoding at all.
EchoKV repurposes positional pseudo-queries for a distinct role: initializing
the first feedback-compatible workspace before any model-generated query
exists. DapQ addresses
where an eviction query should look; EchoKV addresses how to construct the
first workspace that makes subsequent generated queries reliable. The latter
is coupled with feedback-driven workspace repair, which iteratively corrects
the workspace using model-generated queries once content-level evidence about
the answer becomes available.

\paragraph{Long-context retrieval and memory.}
Memory-augmented models extend the effective context through retrieved or
recurrent state, including kNN memory in Memorizing
Transformers~\citep{wu2022memorizing}, recurrent memory tokens in
RMT~\citep{bulatov2022rmt}, and tiered virtual context management in
MemGPT~\citep{packer2023memgpt}. These approaches change the memory
architecture or introduce an external memory-management policy. EchoKV instead
operates on the KV cache of an unmodified decoder at inference time. This makes
long-context RAG and agent memory motivating application settings, not direct
evaluation claims in this paper.

Iterative inference methods, such as speculative decoding and Self-Refine
\citep{leviathan2023fast,madaan2023selfrefine}, adapt computation using
intermediate outputs; EchoKV instead applies feedback to the eviction workspace.

\section{Observations}
\label{sec:obs}

This section measures a query-fidelity pattern that motivates EchoKV's
feedback-driven workspace repair. Let $\mathbf{q}^{*,\ell,g}_k$ denote the
step-$k$ query for layer $\ell$ and KV group $g$, obtained by greedy decoding
on the uncompressed KV cache. Let $\mathcal{I}^{\ell,g}_B(\mathbf{Q})$ be the
top-$B$ token indices selected by query set $\mathbf{Q}$ under the scoring
rule in Eq.~(3). \emph{KV overlap} is
\begin{equation}
\operatorname{Overlap}_k = \frac{1}{|\mathcal{L}||\mathcal{G}|}
\sum_{\ell,g}\frac{\left|\mathcal{I}^{\ell,g}_B(\hat{\mathbf{Q}})
\cap\mathcal{I}^{\ell,g}_B(\mathbf{q}^{*,\ell,g}_k)\right|}{B}.
\end{equation}
where $\mathcal{L}$ and $\mathcal{G}$ are the layer and KV-group sets.
An overlap of $1$ means exact agreement on the retained token indices.

\subsection{Greedy-lookahead fidelity decays with step count}
\label{sec:obs3}

LAQ generates $T$ greedy tokens on a frozen workspace $\mathcal{W}_1$ and
uses all $T$ queries for final eviction.
If every step contributed equal fidelity to $\mathbf{q}^*$, nothing would
be gained by refreshing the workspace mid-lookahead:
one-shot on the best available $\mathcal{W}_1$ would already be optimal.

Table~\ref{tab:decay} shows the opposite. On a frozen workspace, KV overlap
with $\mathbf{q}^*_k$ \emph{falls} as $k$ grows. This decline is consistent
with error accumulation along the query trajectory, although it does not by
itself identify the source of the drift.
Overlap drops from $0.92$ at $k{=}1$ to $0.77$ at $k{=}8$ (a $16\%$ relative
decline), with the steepest degradation occurring in the second half
($0.88 \to 0.83 \to 0.77$ for steps 3, 6, 8).
The \textbf{Refreshed} row shows that rebuilding the workspace at $k{=}T/2$
using round-1 queries halts the decay: overlap recovers in the second half,
reaching $0.85$ at $k{=}6$ and $0.80$ at $k{=}8$ (versus $0.83$ and $0.77$
for the frozen baseline)---a relative improvement of $+2.4\%$ and $+3.9\%$
respectively.
The joint $[\mathbf{Q}_1;\mathbf{Q}_2]$ pool from the refreshed path therefore
better approximates the oracle's selection behavior in this measurement.

\begin{table}[t]
\centering
\small
\setlength{\tabcolsep}{5pt}
\resizebox{\columnwidth}{!}{%
\begin{tabular}{lcccc}
\toprule
 & \multicolumn{4}{c}{\textbf{KV overlap at step} $k$} \\
\cmidrule(lr){2-5}
\textbf{Workspace strategy} & $k{=}1$ & $k{=}3$ & $k{=}6$ & $k{=}8$ \\
\midrule
Frozen ($T$ steps on $\mathcal{W}_1$) & 0.92 & 0.88 & 0.83 & 0.77 \\
\textbf{Refreshed} at $k{=}T/2$ (ours) & 0.92 & 0.88 & \textbf{0.85} & \textbf{0.80} \\
\bottomrule
\end{tabular}
}
\caption{KV overlap with the true step-$k$ query $\mathbf{q}^*_k$ (budget
$B{=}128$, Mistral-7B-Instruct-v0.2, $n{=}20$ samples from HotpotQA, Qasper,
PassageRetrieval-en, and NarrativeQA, averaged over all layers).
Overlap measures the fraction of top-$B$ tokens selected by a workspace-based
pseudo-query that match the oracle selection, where $1.0$ indicates perfect
agreement.
The frozen baseline shows monotonic degradation; refreshing at the midpoint
recovers $+2.4\%$ and $+3.9\%$ relative overlap
at steps 6 and 8.}
\label{tab:decay}
\end{table}
This pattern provides empirical evidence consistent with the proposed
error-propagation mechanism: under a frozen workspace, query--KV overlap
declines across lookahead steps, while mid-point refreshing partially recovers
it. This observation does not isolate the source of the drift, but motivates
the $T/2{+}T/2$ recovery schedule at equal forward-pass count.

\section{Method}
\label{sec:method}

\begin{figure*}[t]
\centering
\vspace{0.3cm}
\includefigure[width=.9\textwidth]{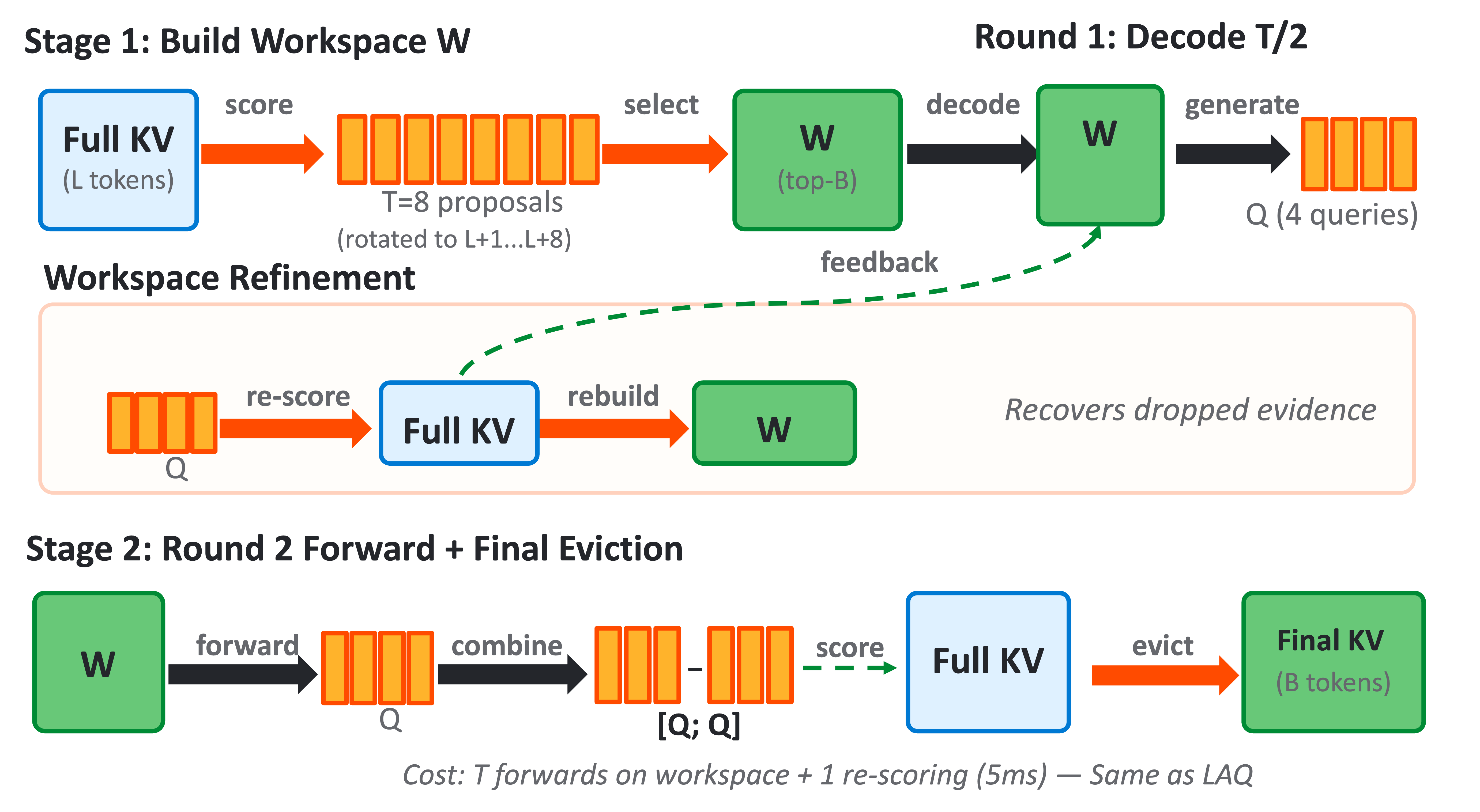}
\vspace{0.2cm}
\caption{EchoKV overview. \textbf{Bootstrap} builds a feedback-compatible
workspace $\mathcal{W}_1$ using future-position proposals (queries rotated to
future decode positions),
then decodes $T/2$ tokens to collect $\mathbf{Q}_1$.
\textbf{Feedback-driven repair:} $\mathbf{Q}_1$ re-scores the full prompt KV to rebuild
$\mathcal{W}_2$.
\textbf{Round 2:} decodes another $T/2$ tokens on $\mathcal{W}_2$ to collect
$\mathbf{Q}_2$; the joint pool $[\mathbf{Q}_1;\mathbf{Q}_2]$ drives final
eviction from $L$ tokens to budget $B$.
Total cost: $T$ forwards on the workspace plus one additional re-scoring pass.}
\label{fig:overview}
\end{figure*}

\subsection{Stage-1: Feedback-Compatible Workspace Bootstrapping}

\paragraph{The workspace selection problem.}
The bootstrap workspace $\mathcal{W}_1$ must be selected at prefill---before any
decode queries are observed---yet it determines which prompt tokens remain
accessible during all $T$ lookahead steps. Its purpose is not to make the final
eviction decision, but to preserve enough evidence for the first generated
queries to support reliable subsequent correction.
Prior work (SnapKV, LAQ) scores the full prompt KV cache using the last $w$
prompt queries, issued at positions $[L{-}w{+}1, L]$.
Under RoPE, these queries assign exponentially decaying attention weights to
earlier tokens purely due to relative-distance penalties, systematically
under-selecting evidence from the prompt's first half even when that evidence
is critical for future decode queries at positions ${\geq}L{+}1$.
We address this with \emph{position-diverse} proposals that approximate the
union of attention patterns across all $T$ decode steps.

\paragraph{Position-aware proposal generation.}
Our construction uses the positional pseudo-query mechanism of
DapQ~\citep{dapq2026}: because the RoPE phase
of a query determines the relative distances at which it attends most strongly,
a query rotated to a future position behaves like a query \emph{issued} at that
position, even though its content is inherited from the prompt.
DapQ uses this to build a decoding-aligned observation window for eviction; we
repurpose it to bootstrap the greedy workspace, where no decoded token is yet
available and position is the only reliable signal. This bootstrap is not a
final eviction score: it establishes the observation substrate on which
model-generated feedback can subsequently be obtained.

Let $L$ be the prompt length, $\mathcal{K}^{\ell,g}=
\{\mathbf{k}^{\ell,g}_i,\mathbf{v}^{\ell,g}_i\}_{i=1}^{L}$ the KV cache for
layer $\ell$ and KV group $g$, and $d_k$ the key dimension. We use $S$ semantic
anchors per layer; in all experiments $S{=}1$, namely the final prompt query.
For anchor $s$, let $\widetilde{\mathbf{q}}^{\ell,g}_s\in\mathbb{R}^{d_k}$ be
its pre-RoPE vector. $\mathbf{R}(p)$ denotes the RoPE rotation at absolute
position $p$. Rotating each anchor to the $T$ future positions yields
$P=S\,T$ proposals:
\begin{equation}
    \mathbf{q}^{\ell,g}_{s,p} =
    \mathbf{R}(p)\,\widetilde{\mathbf{q}}^{\ell,g}_s.
\end{equation}
Here $s\in[S]$ and $p\in\{L+1,\dots,L+T\}$.
For any query set $\mathbf{Q}^{\ell,g}$, we score prompt token $i$ by
accumulating its attention probability across queries:
\begin{equation}
    s_i^{\ell,g}(\mathbf{Q}) =
    \sum_{\mathbf{q}\in\mathbf{Q}^{\ell,g}}
    \frac{\exp(\mathbf{q}^{\top}\mathbf{k}^{\ell,g}_i/\sqrt{d_k})}
    {\sum_{j=1}^{L}\exp(\mathbf{q}^{\top}\mathbf{k}^{\ell,g}_j/\sqrt{d_k})}.
\end{equation}
Here, $\operatorname{TopK}_{B'}$ returns the indices of the $B'$ largest
entries; thus $\mathcal{I}^{\ell,g}_{B'}(\mathbf{Q})=
\operatorname{TopK}_{B'}(s^{\ell,g}(\mathbf{Q}))$ and the bootstrap workspace
is $\mathcal{W}_1^{\ell,g}=\mathcal{I}^{\ell,g}_{B_1}(\mathbf{Q}_{\mathrm{prop}})$.
We evaluate the softmax in chunks over $L$ tokens. For grouped-query attention,
queries are mapped to their shared KV group before applying Eq.~(3); thus all
selection is performed per layer and KV group.

\subsection{Final Query Selection: Why Decode-Generated Queries Dominate}
\label{sec:why_iterative}

Stage~1 selects a workspace; final eviction must make the precise choice of
which $B$ tokens survive. A natural cheap alternative is to score with a
\emph{real} observation window---the last $w$ prompt queries, as SnapKV and its
descendants do. We deliberately do not, for four reasons.

\paragraph{(1) A real window has the wrong RoPE phase.}
The queries that will actually read the surviving cache are issued at absolute
positions $>L$, whereas a real window lives at positions
$[L{-}w{+}1, L]$. Under RoPE, the score $\langle \mathbf{q}_p, \mathbf{k}_i
\rangle$ depends on the relative offset $p-i$, so shifting the query position
reorders which keys rank highest. This is precisely DapQ's observation, and it
is why our stage~1 rotates proposals forward instead of reading the window
directly. A real window cannot be repositioned without ceasing to be real.

\paragraph{(2) A real decode window is not obtainable at prefill.}
One could instead observe \emph{oracle} decode queries by generating the first
few tokens and recording them on the uncompressed cache. Waiting
for a genuine observation window costs $T$ full-cache forward passes at peak
memory $O(L)$ per layer, which forfeits the memory saving that motivates
eviction. Generated pseudo queries run on the $B_1$-sized workspace instead,
so the cost is $O(T\,B_1)$ with $B_1 \ll L$.

\paragraph{(3) Generated queries match the decode distribution in content as
well as position.}
\begin{table}[t]
\centering
\small
\resizebox{0.8\columnwidth}{!}{%
\begin{tabular}{lccc}
\toprule
\textbf{Layer} & \textbf{Locked} & \textbf{Linear} & \textbf{Prompt} \\
\midrule
0  & 0.804 & 0.853 & 0.927 \\
8  & 0.472 & 0.430 & 0.631 \\
16 & 0.275 & 0.334 & 0.599 \\
24 & 0.447 & 0.368 & 0.541 \\
31 & 0.325 & 0.328 & 0.427 \\
\bottomrule
\end{tabular}
}
\caption{KV overlap at step $k{=}3$ (budget 128) for position-only
approximation strategies against the true decode query.
\emph{Prompt}: rotate the last prompt query to future positions.
\emph{Locked}: rotate a real anchor query from a compressed workspace.
\emph{Linear}: linearly extrapolate query content.}
\label{tab:trajectory}
\end{table}

A rotated proposal is position-correct but content-stale: it still carries the
semantics of the prompt's last token. Once a workspace exists, the model can
be run autoregressively on it, and the resulting queries are produced by the
same network, in the same decode regime, at the correct future positions---so
they match the target distribution along both axes. Table~\ref{tab:trajectory}
quantifies the gap that remains for position-only approximations: KV overlap
with the true step-3 query falls to $0.427$ at layer~31, indicating that
position alone is insufficient for final selection in deeper layers.

\paragraph{(4) Only generated queries permit error correction.}
A real window is a single fixed measurement: if the evidence it points to is
incomplete, nothing downstream can detect this. Generated queries provide a
decode-conditioned signal for revisiting stage-1 choices. This is what makes
iteration meaningful: re-scoring the \emph{full} cache with round-1 queries
can recover evidence the initial workspace dropped. A fixed observation window
offers no such feedback signal, and neither does one-shot lookahead on a frozen
workspace.

Points (3) and (4) together imply the specific scheme below: generate, use the
generated queries to rebuild the workspace, then generate again.

\subsection{Feedback-Driven Workspace Repair}

Standard greedy lookahead runs all $T$ steps on a fixed workspace
$\mathcal{W}_1$, accumulating queries $\mathbf{Q}$ that drive final eviction
from the full cache to the final budget $B$. If $\mathcal{W}_1$ omits useful
tokens, later generated queries may be less faithful to the uncompressed-cache
trajectory.

We break this compounding via feedback-driven workspace repair.
The first $T/2$ greedy steps produce queries $\mathbf{Q}_1$ that reveal
which prompt regions the model attempts to attend to, even on incomplete
$\mathcal{W}_1$.
We use $\mathbf{Q}_1$ as a \emph{diagnostic signal} to re-score the
\emph{full} prompt KV cache---still resident at prefill---and rebuild the
workspace to $\mathcal{W}_2$ before continuing.
The second $T/2$ steps then run on $\mathcal{W}_2$, producing queries
$\mathbf{Q}_2$ from the refreshed workspace. The joint pool
$[\mathbf{Q}_1;\mathbf{Q}_2]$ drives final eviction.

\paragraph{Two-round decode scheme.}
Concretely:
\begin{enumerate}[nosep]
    \item \textbf{Round 1.} Decode $T/2$ tokens greedily on $\mathcal{W}_1$
          (size $B_1$), starting from the prompt's boundary token.
          Accumulate per-layer queries $\mathbf{Q}_1$.
    \item \textbf{Rebuild workspace.} Score the full prompt KV cache
          (size $L$) with $\mathbf{Q}_1$ under the same attention-pooling
          rule in Eq.~(2). Select top-$B_1$ tokens per layer and KV group
          $\to \mathcal{W}_2$.
    \item \textbf{Round 2.} Continue greedy decode for the remaining $T/2$
          tokens on $\mathcal{W}_2$, starting from the token generated at
          step $T/2$. Accumulate queries $\mathbf{Q}_2$.
\end{enumerate}

Total forward passes remain $T$ (identical to LAQ).
The workspace rebuild is a scoring operation over existing tensors. Its cost
scales linearly with prompt length and with the $T/2$ round-1 queries per
layer; it requires no additional transformer calls.
Table~\ref{tab:decay} provides supporting evidence: KV overlap with oracle queries
degrades monotonically from $0.92$ to $0.77$ on a frozen workspace but
recovers to $0.85$/$0.80$ at steps 6 and 8 after the midpoint refresh,
consistent with the value of refreshing the workspace mid-lookahead.
The recovery step is one additional KV-scoring pass; its relative cost grows
with context length, as discussed below. Final eviction then uses
$[\mathbf{Q}_1; \mathbf{Q}_2]$ jointly to
select the final $B$ tokens from the full KV cache.

\begin{algorithm}[t]
\caption{EchoKV self-corrective eviction at prefill (per sample)}
\label{alg:paipq}
\begin{algorithmic}[1]
\Require Prompt, budget $B$, $T$, workspace capacity $B_1$
\State Run full prefill; store full KV $\mathcal{K}$
\For{each layer $\ell$}
    \State Build $T$ proposals $\{\mathbf{R}(p)\widetilde{\mathbf{q}}^\ell\}$ via Eq.~(1)
    \State Score $\mathcal{K}^\ell$ with proposals; keep top-$B_1$ $\to \mathcal{W}_1^\ell$
\EndFor
\State Run $T/2$ greedy steps on $\mathcal{W}_1$; collect $\mathbf{Q}_1$
\For{each layer $\ell$}
    \State Re-score $\mathcal{K}^\ell$ with $\mathbf{Q}_1^\ell$; keep top-$B_1$ $\to \mathcal{W}_2^\ell$
\EndFor
\State Run $T/2$ greedy steps on $\mathcal{W}_2$; collect $\mathbf{Q}_2$
\State Re-score $\mathcal{K}$ with $[\mathbf{Q}_1;\mathbf{Q}_2]$; keep top-$B$ $\to$ final cache
\end{algorithmic}
\end{algorithm}
\begin{table*}[t]
\centering
\tiny
\setlength{\tabcolsep}{3pt}
\resizebox{\textwidth}{!}{%
\begin{tabular}{l c ccc ccc ccc ccc cc cc c}
\toprule
 & \textbf{Budget} & \multicolumn{3}{c}{\textbf{Single-Doc}}
 & \multicolumn{3}{c}{\textbf{Multi-Doc}}
 & \multicolumn{3}{c}{\textbf{Summ}}
 & \multicolumn{3}{c}{\textbf{Few-shot}}
 & \multicolumn{2}{c}{\textbf{Synth}}
 & \multicolumn{2}{c}{\textbf{Code}}
 & \\
\cmidrule(lr){3-5}\cmidrule(lr){6-8}\cmidrule(lr){9-11}
\cmidrule(lr){12-14}\cmidrule(lr){15-16}\cmidrule(lr){17-18}
\textbf{Method}
 & & Nrtv & Qas & MF
 & Hot & 2Wiki & Mus
 & Gov & QM & MN
 & TREC & TQA & SAM
 & PC & PR
 & Lcc & RB
 & \textbf{Avg} \\
\midrule
FullKV & -- & 26.90 & 33.14 & 49.26 & 42.77 & 27.35 & 18.77 & 32.95 & 24.21 & 27.13 & 71.00 & 86.23 & 42.72 & 2.75 & 86.98 & 57.12 & 54.51 & 42.74 \\
\midrule
StreamingLLM & 128 & 17.12 & 13.43 & 27.31 & 29.54 & 21.91 & 11.94 & 15.61 & 19.18 & 17.72 & 44.00 & 79.92 & 37.37 & \textbf{3.50} & 23.77 & 51.44 & 45.95 & 28.73  \\
H$_2$O & 128 & 21.62 & 21.34 & 38.61 & 30.46 & 20.38 & 12.20 & 20.59 & \underline{22.51} & 22.03 & 39.00 & 82.33 & 40.64 & \underline{3.06} & \textbf{80.56} & 51.17 & 48.46 & 34.69  \\
SnapKV & 128 & 19.71 & 21.13 & 42.75 & 36.45 & 22.36 & 15.76 & 19.05 & 21.81 & 21.36 & 47.50 & 84.15 & 40.29 & 2.41 & 68.26 & 52.26 & 48.75 & 35.25  \\
PyramidKV & 128 & 21.98 & 22.78 & 43.78 & 32.30 & 22.31 & 15.81 & 20.41 & 21.82 & 21.23 & \underline{66.00} & 83.51 & 39.83 & 2.99 & 65.81 & 51.61 & 46.42 & 36.16  \\
LAQ ($T{=}8$) & 128 & \underline{24.94} & \textbf{27.77} & \underline{45.43} & \underline{40.35} & \textbf{25.25} & \textbf{17.91} & \textbf{22.06} & \textbf{22.68} & \underline{22.50} & \textbf{69.50} & \textbf{86.36} & \underline{41.16} & 1.49 & 76.85 & \underline{53.31} & \textbf{51.02} & \underline{39.29}  \\
EchoKV (ours) & 128 & \textbf{26.11} & \underline{27.51} & \textbf{46.45} & \textbf{41.89} & \underline{24.85} & \underline{17.59} & \underline{21.39} & 22.43 & \textbf{22.56} & \textbf{69.50} & \underline{85.86} & \textbf{41.44} & 2.97 & \underline{78.21} & \textbf{53.45} & \underline{50.52} & \textbf{39.55}  \\
\midrule
StreamingLLM & 256 & 19.22 & 15.08 & 28.01 & 30.40 & 22.02 & 11.21 & 18.09 & 19.59 & 20.09 & 51.00 & 80.71 & 39.89 & 2.96 & 18.40 & 54.30 & 47.38 & 29.90  \\
H$_2$O & 256 & 21.42 & 23.04 & 42.60 & 30.75 & 22.42 & 13.82 & 22.35 & 22.54 & 23.12 & 40.50 & 83.78 & 40.73 & \textbf{3.51} & \textbf{85.85} & 53.18 & 49.95 & 36.22  \\
SnapKV & 256 & 22.44 & 24.07 & \textbf{48.01} & 38.66 & 22.66 & 15.59 & 21.83 & \underline{23.23} & 22.94 & 61.50 & 85.45 & 41.32 & 3.13 & \underline{85.79} & \textbf{55.11} & 51.73 & 38.97  \\
PyramidKV & 256 & 21.69 & 25.18 & 47.61 & 38.77 & \underline{26.12} & 15.23 & 22.52 & 22.52 & 22.59 & 68.00 & 84.27 & 42.10 & \underline{3.43} & 76.60 & 53.08 & 48.40 & 38.63  \\
LAQ ($T{=}8$) & 256 & \underline{24.68} & \underline{29.25} & \underline{48.00} & \underline{40.56} & 26.01 & \underline{18.24} & \textbf{24.04} & 22.96 & \textbf{23.80} & \underline{70.00} & \underline{85.81} & \underline{42.52} & 2.01 & 82.34 & \underline{54.96} & \textbf{53.00} & \underline{40.51}  \\
EchoKV (ours) & 256 & \textbf{25.39} & \textbf{29.74} & 47.97 & \textbf{41.86} & \textbf{26.75} & \textbf{18.71} & \underline{23.61} & \textbf{23.51} & \underline{23.72} & \textbf{70.50} & \textbf{86.46} & \textbf{43.18} & 3.28 & 83.64 & 54.92 & \underline{52.27} & \textbf{40.97}  \\
\midrule
StreamingLLM & 512 & 20.91 & 16.47 & 30.56 & 29.62 & 22.16 & 11.02 & 21.51 & 20.10 & 23.03 & 61.50 & 81.86 & 41.66 & 2.84 & 18.57 & 55.27 & 49.07 & 31.63  \\
H$_2$O & 512 & 21.72 & 25.54 & 44.72 & 32.39 & 23.16 & 14.75 & 23.61 & 23.03 & 24.58 & 42.00 & 85.22 & 41.61 & \textbf{3.42} & \underline{86.45} & 54.82 & 51.11 & 37.38  \\
SnapKV & 512 & 24.14 & 28.11 & \underline{48.78} & 39.49 & 25.09 & 17.40 & 23.67 & 23.18 & 24.60 & 67.00 & 85.88 & 41.39 & 2.78 & \textbf{86.56} & 56.61 & \textbf{53.47} & 40.51  \\
PyramidKV & 512 & 22.99 & 28.74 & 48.45 & 39.73 & 25.74 & 16.58 & 24.48 & 23.40 & 24.52 & \underline{70.00} & 85.99 & \underline{42.40} & \underline{3.32} & 81.63 & 55.93 & 52.38 & 40.39  \\
LAQ ($T{=}8$) & 512 & \underline{24.65} & \textbf{31.21} & \textbf{49.15} & \underline{39.90} & \textbf{27.18} & \underline{18.38} & \textbf{25.55} & \textbf{23.91} & \underline{24.87} & \textbf{71.00} & \underline{86.33} & 42.14 & 1.87 & 86.41 & \textbf{56.84} & \underline{53.08} & \underline{41.40}  \\
EchoKV (ours) & 512 & \textbf{25.52} & \underline{30.69} & 48.73 & \textbf{41.93} & \underline{26.65} & \textbf{19.33} & \underline{25.48} & \underline{23.64} & \textbf{25.29} & \textbf{71.00} & \textbf{86.43} & \textbf{42.43} & 2.75 & 86.35 & \underline{56.64} & 52.99 & \textbf{41.62}  \\
\bottomrule
\end{tabular}
}
\caption{LongBench results on Mistral-7B-Instruct-v0.2 across budgets $B \in \{128, 256, 512\}$.
Column groups: Single-Doc QA (Nrtv=NarrativeQA, Qas=Qasper, MF=MultifieldQA-en); Multi-Doc QA (Hot=HotpotQA, 2Wiki=2WikiMQA, Mus=Musique);
Summ (Gov=GovReport, QM=QMSum, MN=MultiNews); Few-shot (TREC, TQA=TriviaQA, SAM=SAMSum);
Synthetic (PC=PassageCount, PR=PassageRetrieval-en); Code (Lcc, RB=RepoBench-P).
Bold and underline denote the best and second-best compression methods, respectively, within each budget (excluding FullKV).}
\label{tab:mistral_all}
\end{table*}

\begin{table*}[t]
\centering
\tiny
\setlength{\tabcolsep}{3pt}
\resizebox{\textwidth}{!}{%
\begin{tabular}{l c ccc ccc ccc ccc cc cc c}
\toprule
 & \textbf{Budget} & \multicolumn{3}{c}{\textbf{Single-Doc}}
 & \multicolumn{3}{c}{\textbf{Multi-Doc}}
 & \multicolumn{3}{c}{\textbf{Summ}}
 & \multicolumn{3}{c}{\textbf{Few-shot}}
 & \multicolumn{2}{c}{\textbf{Synth}}
 & \multicolumn{2}{c}{\textbf{Code}}
 & \\
\cmidrule(lr){3-5}\cmidrule(lr){6-8}\cmidrule(lr){9-11}
\cmidrule(lr){12-14}\cmidrule(lr){15-16}\cmidrule(lr){17-18}
\textbf{Method}
 & & Nrtv & Qas & MF
 & Hot & 2Wiki & Mus
 & Gov & QM & MN
 & TREC & TQA & SAM
 & PC & PR
 & Lcc & RB
 & \textbf{Avg} \\
\midrule
FullKV & -- & 17.07 & 43.76 & 52.61 & 57.70 & 47.13 & 29.85 & 32.00 & 23.61 & 23.95 & 39.50 & 87.65 & 40.08 & 8.50 & 100.00 & 6.62 & 9.79 & 38.74 \\
\midrule
StreamingLLM & 128 & 11.77 & 23.65 & 26.44 & 40.90 & 37.09 & 16.42 & 16.12 & 18.03 & 14.69 & 12.00 & 76.26 & 35.50 & \textbf{8.50} & 24.50 & \textbf{10.26} & \textbf{13.44} & 24.10  \\
H$_2$O & 128 & 16.61 & 29.58 & 37.25 & 51.04 & 42.14 & 20.82 & 20.33 & \textbf{22.05} & 18.07 & 18.50 & 80.85 & \textbf{39.04} & \textbf{8.50} & \textbf{98.00} & 6.29 & 9.77 & 32.43  \\
SnapKV & 128 & 16.23 & 32.72 & 45.14 & 52.95 & 44.10 & 24.16 & 19.27 & 20.81 & 17.27 & 21.50 & 85.83 & 37.75 & \textbf{8.50} & 95.00 & \underline{7.26} & \underline{11.28} & 33.74  \\
PyramidKV & 128 & 15.40 & 30.67 & 44.89 & 51.54 & 41.22 & 25.62 & 19.51 & 19.71 & 16.69 & 25.75 & 84.22 & 34.08 & \textbf{8.50} & 96.00 & 5.29 & 6.32 & 32.84  \\
LAQ ($T{=}8$) & 128 & \textbf{17.98} & \underline{39.17} & \underline{48.85} & \underline{56.76} & \underline{44.96} & \underline{28.96} & \underline{22.39} & 20.66 & \textbf{18.77} & \underline{35.00} & \textbf{87.78} & 37.75 & \textbf{8.50} & \textbf{98.00} & 5.82 & 9.35 & \underline{36.29}  \\
EchoKV (ours) & 128 & \underline{17.27} & \textbf{40.61} & \textbf{49.65} & \textbf{57.10} & \textbf{46.50} & \textbf{29.25} & \textbf{22.41} & \underline{21.78} & \underline{18.72} & \textbf{39.50} & \underline{87.55} & \underline{37.98} & \textbf{8.50} & \underline{97.50} & 5.99 & 9.83 & \textbf{36.88}  \\
\midrule
StreamingLLM & 256 & 11.50 & 24.44 & 26.81 & 41.85 & 37.25 & 17.39 & 18.90 & 18.12 & 17.11 & 17.50 & 80.66 & 37.47 & \textbf{9.00} & 24.50 & \textbf{8.42} & \textbf{13.54} & 25.28  \\
H$_2$O & 256 & 16.05 & 32.25 & 43.63 & 52.73 & 43.40 & 23.98 & 22.66 & \underline{22.30} & 19.60 & 19.50 & 83.55 & \textbf{39.24} & \underline{8.50} & \textbf{99.50} & \underline{6.71} & 9.29 & 33.93  \\
SnapKV & 256 & 16.25 & 36.47 & 50.35 & 55.44 & 44.37 & 26.62 & 21.96 & 21.76 & 19.35 & 31.00 & 86.89 & 38.47 & \underline{8.50} & 98.00 & 6.52 & 9.98 & 35.75  \\
PyramidKV & 256 & 16.81 & 34.46 & 47.17 & 54.72 & 42.74 & 25.86 & 21.45 & 20.39 & 17.61 & 33.75 & 85.76 & 36.93 & \underline{8.50} & 98.50 & 5.15 & 7.21 & 34.81  \\
LAQ ($T{=}8$) & 256 & \underline{17.62} & \underline{41.16} & \textbf{50.84} & \textbf{57.16} & \underline{46.48} & \textbf{29.37} & \textbf{24.23} & 22.17 & \underline{20.30} & \underline{38.50} & \textbf{88.10} & \underline{39.12} & \underline{8.50} & \underline{99.00} & 6.14 & 9.83 & \underline{37.41}  \\
EchoKV (ours) & 256 & \textbf{18.01} & \textbf{41.53} & \underline{50.83} & \underline{56.92} & \textbf{47.20} & \underline{27.79} & \underline{24.12} & \textbf{22.45} & \textbf{20.41} & \textbf{39.00} & \underline{87.80} & 38.91 & \underline{8.50} & \underline{99.00} & 6.26 & \underline{10.44} & \textbf{37.45}  \\
\midrule
StreamingLLM & 512 & 12.56 & 27.00 & 29.75 & 42.48 & 36.44 & 16.15 & 22.30 & 18.68 & 20.01 & 24.00 & 83.99 & 38.09 & \textbf{8.50} & 24.00 & \textbf{8.36} & \textbf{12.54} & 26.55  \\
H$_2$O & 512 & 16.77 & 35.32 & 45.82 & 52.30 & 44.39 & 24.29 & 24.20 & 22.29 & 21.45 & 21.00 & 86.29 & 39.16 & \textbf{8.50} & \textbf{100.00} & 6.82 & 9.09 & 34.86  \\
SnapKV & 512 & \underline{17.63} & 40.73 & 50.71 & 55.63 & 45.18 & 27.85 & 24.41 & \textbf{22.84} & 21.16 & \underline{38.50} & \textbf{88.12} & 39.21 & \textbf{8.50} & \underline{99.50} & 7.09 & \underline{10.16} & 37.33  \\
PyramidKV & 512 & 17.49 & 37.81 & 50.45 & 56.36 & 44.64 & 27.46 & 23.00 & 20.97 & 19.60 & 35.25 & 87.62 & 37.82 & \textbf{8.50} & 99.00 & 5.62 & 7.94 & 36.22  \\
LAQ ($T{=}8$) & 512 & 17.26 & \textbf{42.64} & \textbf{51.73} & \textbf{57.56} & \underline{46.59} & \textbf{29.12} & \textbf{26.57} & \underline{22.67} & \underline{21.86} & \textbf{39.00} & 87.65 & \underline{39.40} & \textbf{8.50} & \textbf{100.00} & 6.52 & 9.91 & \underline{37.94}  \\
EchoKV (ours) & 512 & \textbf{18.50} & \underline{42.33} & \underline{51.65} & \underline{57.19} & \textbf{47.27} & \underline{28.30} & \underline{26.38} & 22.51 & \textbf{21.89} & \textbf{39.00} & \underline{87.76} & \textbf{39.54} & \textbf{8.50} & \textbf{100.00} & \underline{7.16} & 9.88 & \textbf{37.99}  \\
\bottomrule
\end{tabular}
}
\caption{LongBench results on Qwen2.5-7B-Instruct across budgets $B \in \{128, 256, 512\}$.
Same column order as Table~\ref{tab:mistral_all}. Bold and underline denote the best and second-best compression methods, respectively, within each budget (excluding FullKV).}
\label{tab:qwen_all}
\end{table*}

\paragraph{Complexity and overhead.}
EchoKV adds to standard full-cache prefill two costs:
(a)~$T$ greedy decode steps over a $B_1$-sized workspace---cost
$O(T \cdot B_1 \cdot d)$, identical to LAQ, with $B_1 \ll L$; and
(b)~one mid-point re-scoring pass over the full prompt KV using $T/2$ query
vectors per layer. EchoKV, like other post-prefill eviction methods, retains
the full KV cache until final selection; it then reduces the retained decode
cache to $B$ tokens. No additional transformer forward passes are required,
and recovery is a tensor re-scoring operation. EchoKV therefore has
end-to-end latency that depends on prompt length and implementation; we
therefore do not claim latency parity beyond the equal number of workspace
forward passes.

\paragraph{Potential applications.}
Several tasks on which EchoKV improves, including \textsc{HotpotQA} and
\textsc{2WikiMQA}, require connecting facts from non-adjacent context segments.
Exactly this pattern arises in (i)~\emph{Long-Context RAG}, where retrieved
passages are concatenated and the model must locate relevant spans scattered
across the context; (ii)~\emph{multi-document reasoning}, where supporting
evidence appears in different source documents; and
(iii)~\emph{agent long-context memory}, where tool outputs and observations
from many prior turns accumulate in the context window and must be recalled
coherently.
Standard window-biased eviction can drop this dispersed evidence early;
EchoKV's feedback-compatible bootstrap and feedback-driven repair are designed to
preserve it. These are motivating application patterns, rather than claims of
performance on dedicated RAG or agent-memory benchmarks; evaluating them on
such benchmarks is future work.

\section{Experiments}
\label{sec:experiments}

\subsection{Setup}
\paragraph{Model and data.}
We evaluate on two backbones with Flash Attention~2:
Mistral-7B-Instruct-v0.2 and Qwen2.5-7B-Instruct.
Mistral serves as the primary model, matching the setting of
LAQ~\citep{wang2025lookahead} so that our numbers are directly comparable to
published baselines; Qwen2.5-7B-Instruct tests whether the method transfers
across attention configurations (Qwen2.5 uses grouped-query attention with a
different head/KV-group ratio).
Both are evaluated on the 16 English subsets of
LongBench~\citep{bai2023longbench}, spanning single-document QA
(\textsc{NarrativeQA}, \textsc{Qasper}, \textsc{MultifieldQA-en}),
multi-document QA (\textsc{HotpotQA}, \textsc{2WikiMQA}, \textsc{Musique}),
summarization (\textsc{GovReport}, \textsc{QMSum}, \textsc{MultiNews}),
few-shot learning (\textsc{TREC}, \textsc{TriviaQA}, \textsc{SAMSum}),
synthetic retrieval (\textsc{PassageCount},
\textsc{PassageRetrieval-en}), and code completion
(\textsc{Lcc}, \textsc{RepoBench-P}).

\paragraph{Baselines.}
We report FullKV (no eviction) and compare against StreamingLLM,
H$_2$O~\citep{zhang2023h2o}, SnapKV~\citep{li2024snapkv},
PyramidKV~\citep{cai2024pyramidkv}, and LAQ~\citep{wang2025lookahead}
(greedy, $T{=}8$). The component variants are described in
Section~\ref{sec:ablation}.

\paragraph{Evaluation protocol.}
We use the official LongBench prompts and task metrics, greedy generation, and
the same cache budgets for every compression method. Scores are computed over
the 16 English LongBench subsets listed above. Unless stated otherwise, the
analysis experiments use Mistral-7B-Instruct-v0.2 at $B{=}128$; all workspace
selection is performed independently per layer and KV group.

\paragraph{EchoKV settings.}
$S{=}1$ (single semantic anchor), $T{=}8$ lookahead steps (two rounds of 4),
workspace capacity $B_1{=}1024$, \texttt{stage2\_window}{=}0.
KV budgets $B \in \{128, 256, 512\}$ to match prior work.

\subsection{Main Results}

Table~\ref{tab:mistral_all} reports full LongBench scores for Mistral-7B-Instruct-v0.2 across all three budgets ($B \in \{128, 256, 512\}$).
Table~\ref{tab:qwen_all} shows the same for Qwen2.5-7B-Instruct.

EchoKV is competitive with LAQ across budgets and backbones. At $B{=}128$, it
improves the 16-task average over LAQ on Qwen ($36.88$ vs.\ $36.29$) and
Mistral ($39.55$ vs.\ $39.29$). The per-task pattern is more informative than
the average: EchoKV's clearest gains occur on selected dispersed-evidence
tasks, whereas it is not uniformly superior elsewhere.

\paragraph{Budget scaling.}
Tables~\ref{tab:mistral_all} and~\ref{tab:qwen_all} show results across all three
budgets ($B \in \{128, 256, 512\}$) for both models.
Aggregate rankings vary by budget and backbone. EchoKV remains close to the
LAQ baseline on average while delivering notable gains on selected
retrieval-intensive tasks.

\subsection{Cross-Model Generalization}

Tables~\ref{tab:mistral_all} and~\ref{tab:qwen_all} evaluate EchoKV on both
Mistral-7B-Instruct-v0.2 and Qwen2.5-7B-Instruct (which uses grouped-query
attention with a different head/KV-group ratio), across all three budgets
($B \in \{128, 256, 512\}$). Both architectures show that the method remains
competitive overall, while the size and location of task-level gains vary.

\paragraph{Comparison with baselines.}
On Qwen at $B{=}128$, EchoKV improves over LAQ ($T{=}8$) on
\textsc{HotpotQA} ($57.10$ vs.\ $56.76$) and
\textsc{2WikiMQA} ($46.50$ vs.\ $44.96$). Its aggregate advantage is not
uniform across budgets. This distinction is important: EchoKV is intended to
address failure cases associated with dispersed evidence, not to replace every
eviction strategy on every task.

\paragraph{Code-subset anomaly.}
\textsc{Lcc} and \textsc{RB} have unusually low FullKV scores on Qwen
($6.62$ and $9.79$), so the code-subset results should be interpreted with
caution. We do not attribute this pattern to cache eviction without a
separate validation of the evaluation pipeline; conclusions about transfer
should therefore be read together with the Mistral results and non-code Qwen
subsets.
\section{Analysis}
\label{sec:analysis}

\subsection{Targeted Comparison with LAQ}
\label{sec:ablation}

Table~\ref{tab:ablation} compares EchoKV directly with the LAQ baseline on
two evidence-intensive tasks at the most constrained budget.

\begin{table}[t]
\centering
\small
\resizebox{\columnwidth}{!}{%
\begin{tabular}{lccc}
\toprule
\textbf{Method} & \textbf{MultifieldQA} & \textbf{HotpotQA} & \textbf{AVG} \\
\midrule
LAQ ($T{=}8$, baseline) & 45.43 & 40.35 & 39.29 \\
\quad + EchoKV-Stage1 only & 46.15 & \underline{41.24} & 39.38 \\
\quad + Iterative only & \underline{46.27} & 40.98 & \underline{39.42} \\
\textbf{EchoKV} & \textbf{46.45} & \textbf{41.89} & \textbf{39.55} \\
\bottomrule
\end{tabular}
}
\caption{Component ablation on Mistral-7B-Instruct-v0.2 at $B{=}128$.
  Each row adds the indicated EchoKV component to LAQ. Combining both stages
  yields complementary task-level gains: $+1.02$ on MultifieldQA, $+1.54$
  on HotpotQA, and a modest $+0.26$ on the 16-task average over LAQ. Bold
  denotes the best and
  underlining the second-best result in each column.}
\label{tab:ablation}
\end{table}

At this budget, the components show complementary effects. Stage~1 alone gives
the stronger HotpotQA gain, whereas iterative repair alone gives the stronger
MultifieldQA and average gains among the single-component variants. Their
combination improves both evidence-intensive tasks simultaneously, although
the aggregate gain is modest ($+0.26$). This pattern aligns with the roles of
bootstrapping a useful observation workspace and then revising it with
generated-query feedback before final eviction.

\subsection{Evidence Retention Analysis}
\label{sec:evidence-retention}

We measure supporting-evidence retention after refinement: the percentage of
task-provided gold evidence-span tokens retained in the workspace. We map spans
to input-token positions and average per-example retention over layers and KV
groups. Table~\ref{tab:evidence_retention} compares LAQ's workspace (W) with
EchoKV after bootstrapping (W1) and repair (W2), at $B{=}128$ on Mistral.

\begin{table}[t]
\centering
\small
\resizebox{\columnwidth}{!}{%
\begin{tabular}{lccc}
\toprule
\textbf{Task} & \textbf{LAQ W} & \textbf{EchoKV W1} & \textbf{EchoKV W2} \\
\midrule
\textsc{HotpotQA} & 68.4 & \underline{72.1} & \textbf{76.8} \\
\textsc{2WikiMQA} & 66.9 & \underline{71.2} & \textbf{75.4} \\
\textsc{MultifieldQA} & 70.6 & \underline{73.5} & \textbf{77.0} \\
\textsc{Qasper} & 74.2 & \underline{76.0} & \textbf{78.1} \\
\midrule
\textbf{Average} & 70.0 & \underline{73.2} & \textbf{76.8} \\
\bottomrule
\end{tabular}
}
\caption{Evidence-token retention (\%) at $B{=}128$ on Mistral-7B-Instruct-v0.2.
Gold evidence spans are mapped to input-token positions; W, W1, and W2 denote
LAQ's workspace, EchoKV's bootstrap workspace, and its repaired workspace.
Bold denotes the best and underlining the second-best result in each row.}
\label{tab:evidence_retention}
\end{table}

W1 and W2 progressively retain more evidence, including on \textsc{HotpotQA}
and \textsc{MultifieldQA}. The $+3.2$-point W-to-W1 and $+3.6$-point W1-to-W2
average gains exceed downstream task gains, as retained evidence need not
change an answer. This is a mechanism diagnostic, not a causal proof.

\subsection{Discussion}
\label{sec:discussion}

EchoKV is most useful when decoding reconnects distant evidence, as in
\textsc{HotpotQA} and \textsc{2WikiMQA}. The overlap and retention analyses
are consistent with this mechanism, not causal proof. Recency-biased tasks may
already be served by window eviction; RAG, agent-memory, and long-context
latency evaluations remain future work.

\section{Conclusion}

EchoKV bootstraps a workspace with future-position proposals and repairs it
with early generated queries. It is competitive with LAQ on LongBench, with
targeted gains on retrieval-intensive tasks, while requiring full-cache
retention until final selection and one extra re-scoring pass.
The overlap and evidence-retention analyses further show that workspace
refreshing improves later query fidelity and retains more annotated evidence,
which is consistent with the proposed closed-loop correction mechanism.
These findings position EchoKV as a targeted inference-time complement to
window-based eviction: it is most useful when a future query must reconnect
non-local evidence, rather than as a universal replacement for all cache
compression strategies.

\section*{Limitations}

Proposal scoring costs $O(P L)$ per layer and can be non-trivial at long
contexts. We do not isolate random/oracle refreshes or alternative intervals;
\texttt{stage2\_window}$\,{=}\,0$ also provides no recency guarantee.

\bibliography{custom}

\end{document}